\documentclass{article}

\usepackage{microtype}
\usepackage{graphicx}
\usepackage{subfigure}
\usepackage{booktabs}
\usepackage{multirow}
\usepackage{hyperref}
\usepackage{amsmath}
\usepackage{amssymb}
\usepackage{mathtools}
\usepackage{array}
\usepackage{enumitem}
\usepackage[capitalize, noabbrev]{cleveref}
\usepackage{float}
\usepackage{wrapfig}
\usepackage[preprint]{icml2026}

\usepackage{soul, xcolor}

\newcommand{\rev}[1]{#1}

\icmltitlerunning{On-Device Dental Image Understanding via Efficient Multimodal Large Language Models}

\begin{document}
    \twocolumn[ \icmltitle{Pocket-Dentist: On-Device Dental Image Understanding via Efficient Multimodal Large Language Models}

    \icmlsetsymbol{equal}{*}

    \begin{icmlauthorlist}
        \icmlauthor{Kai Bian}{equal,uoa}
        \icmlauthor{Xucheng Guo}{equal,sdu}
        \icmlauthor{Bin Chen}{unimelb}
        \icmlauthor{Lingyan Ruan}{unimelb}
        \icmlauthor{Yiran Shen}{sdu}
        \icmlauthor{Ting Dang}{unimelb}
        \icmlauthor{Hong Jia}{uoa}
    \end{icmlauthorlist}

    \icmlaffiliation{uoa}{The University of Auckland, New Zealand}
    \icmlaffiliation{sdu}{Shandong University, China}
    \icmlaffiliation{unimelb}{The University of Melbourne, Australia}
    \icmlcorrespondingauthor{Hong Jia}{hong.jia@auckland.ac.nz}
    \icmlkeywords{efficient multimodal QA, dental AI, vision-language models, on-device inference, resource-constrained deployment}

    \vskip 0.3in ]

    \printAffiliationsAndNotice{}  

    \begin{abstract}
        Evaluations of dental vision–language models remain fragmented across datasets, task definitions and metrics, and often ignore their computational cost. This limits their widespread deployment for dental screening outside specialist centres, where timely inference, limited hardware, and local handling of patient images are vital for practical, privacy-preserving clinical prescreening. Here we present Pocket-Dentist, an efficiency-aware benchmark for dental multimodal question answering that brings together three datasets spanning $\sim$1,159 patients \rev{(from BRAR and MetaDent)}, five task types and seven metrics. Across typical 14 VLMs, our results reveal an interesting observation: compact VLMs (e.g., 2B-parameter models) \rev{become competitive with much larger VLMs on most metrics after lightweight adaptation} while requiring substantially lower computational costs in dental image understanding. Deployed locally on an iPhone 17 Pro, our finetuned compact VLM Pocket-Dentist-2B processed each sample in 4.31s, reducing latency by 4.9$\times$ and memory use by 2.3$\times$ compared with a 7B baseline. \rev{Our project page is available at} \href{https://2026-icml.github.io/pocket-dentist-icml} {\textcolor{magenta}{\nolinkurl{https://2026-icml.github.io/pocket-dentist-icml}}}.
    \end{abstract}


    \section{Introduction}
    \label{sec:intro}

    \begin{figure*}[t]
        \centering
        \includegraphics[width=\textwidth]{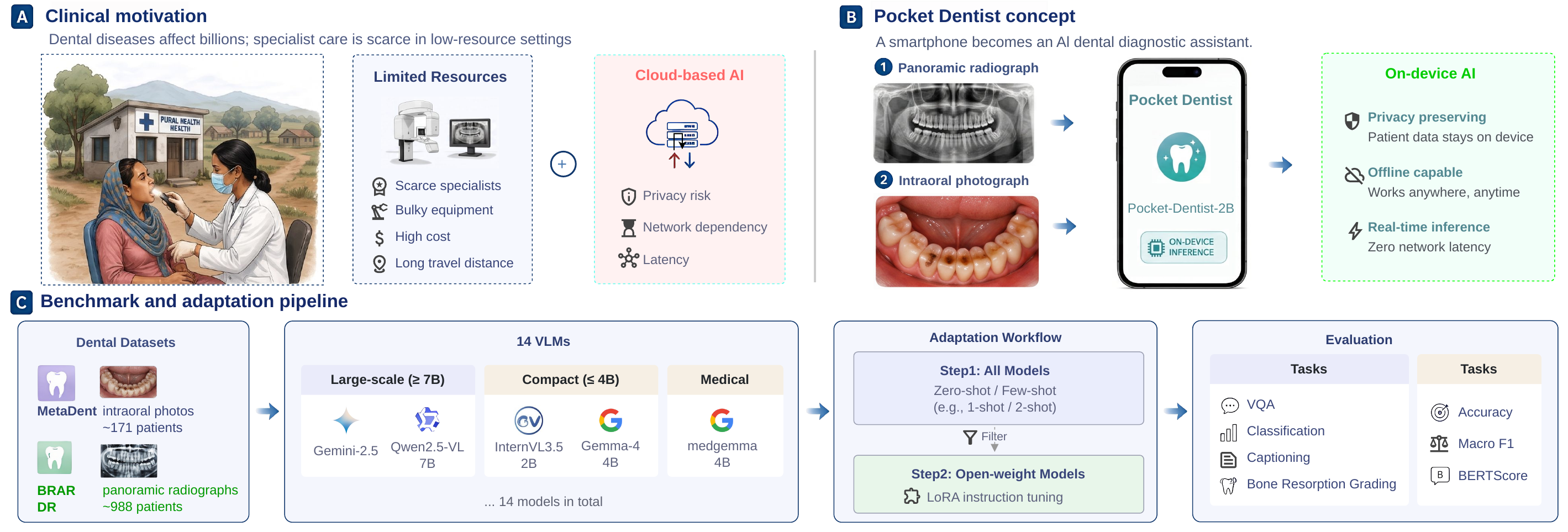}
        \caption{Overview of the Pocket-Dentist pipeline.
        We benchmark 14 VLMs across three dental datasets under zero-shot,
        few-shot, and LoRA settings, identify InternVL3.5-2B as the
        best-performing compact VLM, and deploy it on an iPhone~17~Pro for
        on-device local inference.}
        \label{fig:pipeline}
    \end{figure*}

    Oral diseases are among the most prevalent chronic health conditions globally, affecting approximately 3.7 billion people worldwide~\cite{gbd2025oral}. In remote and underserved communities, patients frequently lack access to even basic oral health screening, let alone specialist diagnosis. This access gap is also an efficiency problem: preliminary screening must be delivered with limited clinical labor, intermittent connectivity, low-cost hardware, and minimal turnaround time. If visual AI models can run directly on widely available mobile devices, dental screening could reach communities that currently lack access to specialist care.

    In practice, dental image understanding is not a single-image, single-label problem. Dental care relies on heterogeneous visual evidence, including panoramic radiographs and intraoral photographs, and the relevant outputs range from short answers and diagnostic categories to captions and structured reports. Yet current dental AI evaluation largely studies these pieces in isolation: datasets are usually modality-specific, task definitions are not standardized across resources, and benchmark results rarely indicate whether a model that performs well in one setting remains reliable under another~\cite{uribe2024datasets,li2026metadent}. This fragmentation becomes more consequential when efficiency is considered. A model intended for practical dental QA must be evaluated not only for answer quality, but also for adaptation cost, structured-output reliability, memory use, latency, and independence from cloud inference. Existing work, therefore, leaves open a central question: \textit{whether efficient multimodal QA models can achieve reliable dental image understanding across diverse modalities and tasks while remaining feasible for resource-constrained mobile deployment.}

    Efficiency is therefore a deployment requirement rather than a
    secondary optimization for this application. In underserved and rural
    screening scenarios, reliable connectivity cannot be assumed, and
    cloud-based multimodal QA requires transmitting patient dental images
    to remote servers, raising both availability and privacy concerns.
    On-device inference addresses these constraints by keeping data local
    and enabling offline, low-latency screening, but it also introduces
    strict memory, compute, and thermal budgets. These constraints
    motivate compact VLMs and make it necessary to evaluate dental QA models
    not only by answer quality, but also by whether they can operate within
    the resource envelope of a mobile device.

    To address these gaps, we introduce Pocket-Dentist (Figure~\ref{fig:pipeline}), an efficiency-aware multimodal QA benchmark and evaluation framework for dental VLMs. Pocket-Dentist formulates dental image understanding as a unified multimodal QA problem in which models are evaluated across heterogeneous imaging modalities, multiple clinical question types, structured output requirements, adaptation regimes, and mobile deployment constraints. We instantiate this setting by curating three heterogeneous dental datasets into standardized prompt--response QA pairs through task-specific reformulation, LLM-assisted conversion, and manual quality review, yielding a unified evaluation setup across all tasks. We evaluate 14 representative VLMs under zero-shot, few-shot, and LoRA instruction-tuning settings, characterizing the accuracy--efficiency trade-off from 1B to 32B parameters, and we deploy a LoRA-tuned compact model on an iPhone~17~Pro with on-device inference.

    Our main contributions are:
    \begin{itemize}[nosep]
        \item We curate Pocket-Dentist, a multimodal dental QA benchmark that unifies three datasets across panoramic radiographs and intraoral photographs, covering $\sim$1{,}159 patients \rev{(from BRAR and MetaDent)}, five clinical question types, and seven evaluation metrics.
        \item We propose an efficiency-aware evaluation framework that unifies heterogeneous dental resources through shared task definitions, structured output schemas, and seven complementary metrics spanning answer quality, adaptation cost, output reliability, memory usage, and inference latency.
        \item We conduct an efficiency-aware evaluation of 14~VLMs, characterizing the accuracy--efficiency trade-off across model scales. Under a uniform LoRA budget, compact adapted VLMs become competitive with larger open-weight models, and we deploy Pocket-Dentist-2B locally on an iPhone~17~Pro with 4.31s per-sample latency, a 4.9$\times$ reduction relative to the 7B baseline.
    \end{itemize}


    \section{Related Work}
    \label{sec:related}

    \textbf{Medical VLMs.} Medical vision--language models have progressed from task-specific medical VQA datasets and benchmarks for medical visual question answering~\cite{he2020pathvqa, zhang2023pmcvqa} to more general-purpose medical vision--language models such as LLaVA-Med~\cite{li2023llavamed} and Med-Flamingo~\cite{moor2023medflamingo} that leverage large-scale medical image--text pretraining. Recent work in medical AI has also emphasized clinical alignment, safety, and human evaluation as prerequisites for reliable deployment~\cite{singhal2023clinical}. However, medical VLM evaluation remains concentrated on radiology and pathology. Dentistry differs from these domains in image acquisition, anatomical structure, annotation conventions, and potential mobile screening workflows, yet it has received limited attention as a multimodal VLM benchmark domain. As a result, it remains largely unclear how current VLMs handle dental imaging tasks or whether performance patterns observed in radiology and pathology benchmarks generalize to dental image understanding.

    \textbf{Dental AI benchmarks.} In dentistry, AI research has largely focused on task-specific models for caries detection, while recent work has highlighted the value of multimodal integration for diagnostic support~\cite{zhang2024caries, uddin2025cariesreview}. MetaDent~\cite{li2026metadent} introduced a dental imaging benchmark covering VQA, captioning, and classification from expert-annotated intraoral photographs, while BRAR~\cite{xia2025brar} contributed panoramic radiograph annotations for periodontal bone loss grading. These resources are valuable, but they do not by themselves define an efficiency-aware dental QA benchmark. Without such a benchmark, there is no systematic way to determine whether compact VLMs can provide reliable dental QA across diverse tasks, or to identify the accuracy--efficiency frontier that governs practical deployment decisions. More importantly, prior work has not curated heterogeneous dental datasets into a single VLM evaluation setting that jointly covers multiple modalities, task types, adaptation regimes, structured-output reliability, and deployment efficiency.

    \textbf{On-device deployment and efficient QA.} On-device deployment of language models has attracted increasing attention as a route toward privacy-preserving and low-latency healthcare AI~\cite{healthslmbench, lu2025demystifying}. HealthSLM-Bench~\cite{healthslmbench} evaluated text-only small language models on wearable health tasks and assessed smartphone deployment efficiency. Lu et al.~\cite{lu2025demystifying} systematically examined the capabilities and runtime characteristics of small language models for edge deployment. More broadly, the efficient QA community has studied accuracy--efficiency trade-offs in retrieval-augmented and compute-constrained settings, but primarily for text-only models in general domains. These efforts do not address the specific demands of multimodal vision--language QA, which must process high-resolution images alongside text under tight memory and latency budgets. Our work brings this efficiency-aware QA perspective to dental VLMs by jointly evaluating answer quality, token throughput, adaptation behavior, reliability failures, and smartphone inference within the same benchmark pipeline.


    \section{Pocket-Dentist}
    \label{sec:setup}

    \subsection{Benchmark Curation}
    \label{sec:datasets}

    We evaluate three dental benchmarks spanning panoramic radiography and intraoral photography, covering five clinical tasks. Table~\ref{tab:datasets} summarizes the dataset characteristics. The curated corpus is reformulated into standardized VLM tasks covering seven evaluation metrics. For cross-model comparison, each task is assigned a single primary metric chosen to reflect clinical priorities. Classification tasks (BRAR, DR, MetaDent) use Macro~F1 rather than accuracy because the label distributions are imbalanced: a model predicting only the majority class would achieve misleadingly high accuracy while failing on clinically important minority conditions such as severe bone resorption or rare pathologies. Captioning uses BERTScore~F1~\cite{zhang2020bertscore} instead of surface-level $n$-gram metrics (BLEU, ROUGE) because clinical descriptions exhibit high lexical variation---semantically equivalent reports may use different terminology, making token-overlap metrics unreliable indicators of diagnostic fidelity. VQA uses exact-match accuracy against clinician-verified reference answers, where incorrect responses (e.g., misidentifying a lesion as normal) could directly delay treatment. Formal metric definitions are provided in Appendix~\ref{app:eval-metrics}. Each dataset uses its original expert-curated train/val/test split; no additional patient-level deduplication was applied.

    \begin{table*}[t]
        \caption{Summary of benchmark datasets and task formulations.}
        \label{tab:datasets}
        \centering
        \small
        \setlength{\tabcolsep}{3pt}
        \begin{tabular}{llllccc}
            \toprule
            Dataset & Modality & Task & Metric
            & Train & Val & Test \\
            \midrule
            BRAR    & Panoramic Radio. & Classification & Macro F1
            & 691  & 148 & 149 \\
            \midrule
            \multirow{3}{*}{MetaDent}
                    & \multirow{3}{*}{Intraoral Photo}
                    & VQA            & Accuracy
            & \multirow{3}{*}{7,238} & \multirow{3}{*}{906} & \multirow{3}{*}{2,301} \\
                    &  & Captioning     & BERTScore F1 &  &  &  \\
                    &  & Classification & Macro F1     &  &  &  \\
            \midrule
            DR      & Panoramic Radio. & Classification & Macro F1
            & 1,075 & 121 & 73 \\
            \bottomrule
        \end{tabular}
    \end{table*}

    BRAR~\cite{xia2025brar} is a multimodal dataset comprising 1,104 panoramic radiographs from 1,104 patients collected at Shanghai Stomatological Hospital between January 1 and March 31, 2025. The cohort includes adults aged 18--75 years. Bone resorption severity is graded into three levels (Grade 1/2/3) using the Bone Resorption Age Ratio (BRAR) index. After excluding radiographs with incomplete metadata, 988 images are used, with a train/val/test split of 691/148/149.

    MetaDent~\cite{li2026metadent} is a large-scale dental image resource with 60{,}669 images from clinical, public, and web sources. For this study, we used its expert-curated subset of 2{,}588 high-quality intraoral photographs (around 171 patients), each annotated with a semi-structured meta-label describing imaging perspective, anatomical focus, and observed abnormalities. The data were split into 7{,}238/906/2{,}301 training/validation/test samples across three tasks (261 test images).

    The Dental Radiography (DR) dataset~\cite{dr2023kaggle} is a
    community-contributed collection of 1{,}269 panoramic dental
    X-rays hosted on Kaggle. Each image is annotated with per-object labels
    across four lesion categories: Cavity, Fillings, Impacted
    Tooth, and Implant. We derive image-level multi-label
    classification targets by extracting the unique set of
    categories present in each image.
    We use 1{,}075/121/73 train/val/test images.
    Because this Kaggle collection does not provide patient-level
    identifiers, the ${\sim}$1{,}159 patient count reported in
    this paper covers only BRAR and MetaDent.

    \subsection{Data Processing and Task Formulation}
    \label{sec:processing}

    We transform all three benchmarks into dataset-specific
    prompt--response pairs to support unified VLM evaluation and
    supervised instruction tuning.

    For BRAR, we formulate a multimodal radiograph-plus-metadata
    classification task: each prompt presents the panoramic image
    together with patient-level metadata (age, gender, missing teeth,
    implants, residual roots, functional tooth count) and requests a
    single severity grade as a JSON object. To ensure reliable
    evaluation, we normalize predictions using a deterministic
    five-level fallback parser
    (see Appendix~\ref{app:prompt-details} for details).

    For MetaDent, we instantiate three task-specific prompt--response
    formats from the expert-authored meta-labels via LLM-assisted
    conversion~\cite{li2026metadent}: VQA (one image--question pair
    per sample), multi-label classification (18 dental condition
    categories), and clinical captioning (free-text description).
    To mitigate the task imbalance in the training and validation splits, we
    subsample VQA to approximately 1.5 pairs per image, yielding
    a balanced task ratio. The test split retains the full set of
    QA pairs per image, since VQA accuracy is aggregated at the
    image level (Section~\ref{sec:results}). The classification set and a random subset
    of VQA pairs were manually reviewed, and uncertain entries
    were excluded.

    For DR, we aggregate per-image CSV annotations to extract
    the unique set of lesion categories present in each image,
    yielding a multi-label classification target. The model is
    prompted to identify which of the four lesion categories
    (Cavity, Fillings, Impacted Tooth, Implant) are present in
    the panoramic radiograph, evaluated via Macro~F1 and Accuracy.

    \subsection{Training and Decoding}
    \label{sec:implementation}

    Let each benchmark task be represented as a set of image--prompt--target triples $\mathcal{D}_t=\{(x_i,p_i,y_i)\}_{i=1}^{n_t}$, where $x_i$ is a dental image, $p_i$ is the structured task prompt, and $y_i$ is the reference output. For a VLM with parameters $\theta$, zero-shot decoding predicts
    \begin{equation}
    \hat{y}^{\mathrm{ZS}}_i
    = \arg\max_y p_{\theta}(y \mid x_i, p_i).
    \end{equation}
    Few-shot evaluation augments the query with a fixed exemplar set $\mathcal{E}_K=\{(x_j,p_j,y_j)\}_{j=1}^{K}$, shared across models for each task:
    \begin{equation}
    \hat{y}^{\mathrm{FS}}_i
    = \arg\max_y p_{\theta}(y \mid \mathcal{E}_K, x_i, p_i).
    \end{equation}
    For LoRA instruction tuning, only the low-rank adapter parameters $\phi$ are optimized while the base model parameters $\theta$ remain frozen:
    \begin{equation}
    \phi^{*}
    = \arg\min_{\phi}
    \sum_{t}\sum_{(x_i,p_i,y_i)\in\mathcal{D}_t}
    -\log p_{\theta+\Delta_{\phi}}(y_i \mid x_i,p_i).
    \end{equation}
    The adapted model then predicts $\hat{y}^{\mathrm{SFT}}_i=\arg\max_y p_{\theta+\Delta_{\phi^{*}}}(y\mid x_i,p_i)$.

    Each model is evaluated under three settings that correspond to different levels of deployment preparation. Under the zero-shot (ZS) setting, models are evaluated without task-specific training using structured prompts and greedy decoding (\texttt{do\_sample=False}) for deterministic, reproducible predictions. Under the few-shot (FS) setting, following Brown et al.~\cite{brown2020fewshot}, we evaluate 1-shot and 2-shot in-context learning with fixed exemplars shared across all models. We report both settings separately in Table~\ref{tab:fs} to avoid post-hoc shot-count selection on the test set. We limit the evaluation to two shots because VLM context windows must accommodate both the exemplar images and the query image; moreover, recent work on health-domain SLMs~\cite{healthslmbench} observed diminishing or negative returns beyond 1--3 shots, a pattern we also observe in Section~\ref{sec:zs}.

    For LoRA instruction tuning (SFT), we apply LoRA~\cite{hu2022lora} with rank $r{=}16$ and $\alpha{=}32$, targeting all linear layers. Training runs for 3 epochs with cosine learning rate scheduling (peak $\text{lr}{=}2{\times}10^{-4}$, warmup ratio 0.1), effective batch size~16, and gradient checkpointing. All task-specific LoRA configurations share identical hyperparameters across all three benchmarks, so differences across models reflect adaptation behavior under a controlled low-cost budget rather than task-specific tuning. All tasks use the standard next-token prediction loss. Inference uses greedy decoding throughout. All experiments are conducted on NVIDIA H100 96\,GB GPUs using PyTorch~2.5.1.


    \section{Experimental Setup}

    \subsection{Models}
    \label{sec:models}

    We benchmark 14 vision--language models in two tiers: (i)~large-scale VLMs ($\geq$7B), including three open-source models (Lingshu-32B, MedMO-8B-Next, Qwen2.5-VL-7B~\cite{qwen25vl}) and two closed-source APIs (Gemini-2.5-Flash, GPT-4o-mini); and (ii)~compact VLMs ($\leq$4B), including seven general-purpose models (Qwen3-VL-4B, gemma-4-E4B-it~\cite{gemma4_model_card, gemma}, InternVL2.5-4B~\cite{internvl}, SmolVLM2-2.2B~\cite{smolvlm}, InternVL3.5-2B, gemma-4-E2B-it, InternVL3.5-1B) and two medical-domain models pre-trained on biomedical corpora (medgemma-4b-it, paligemma2-3b-mix-448). This model pool is chosen to support efficiency-aware comparison across parameter scale, access type, and pretraining source.

    \vspace{-0.3em}

    \subsection{On-Device Deployment Setup}
    \label{sec:deploy-setup}

    To assess systems-level feasibility for on-device deployment, we deploy the top-performing
    LoRA-tuned VLMs on an iPhone~17~Pro (A19~Pro SoC, 12\,GB Unified Memory).
    All models are quantized to GGUF Q4\_K\_M (4-bit) and executed via
    \texttt{llama.cpp}~\cite{gerganov2023llama} mapped to the Metal GPU API through a native Swift
    \texttt{LocalLLMClient} package. This on-device architecture keeps
    all inference local to the device, avoiding the privacy and latency
    concerns inherent in cloud-based
    deployment~\cite{das2025security, lu2025demystifying}. Models are
    sideloaded to the app's Documents directory and loaded into Unified
    Memory for hardware-accelerated inference.

    We deploy three models to quantify the efficiency--capability frontier:
    Pocket-Dentist-2B (InternVL3.5-2B + LoRA, our best overall compact VLM),
    InternVL2.5-4B, and
    Qwen2.5-VL-7B (7B baseline).
    The iPhone~17~Pro was selected because its 12\,GB Unified Memory
    can accommodate the 7B baseline (6.03\,GB). However, Pocket-Dentist-2B itself requires only 2.62\,GB RAM, placing it well within the capability of mid-range
    devices with 4\,GB+ memory (e.g., recent Android phones in the
    \$150--200 range), substantially lowering the hardware barrier
    relative to the test platform used in this study.


    \section{Results}
    \label{sec:results}

    \subsection{Zero-Shot Fragility and Few-Shot Limits}
    \label{sec:zs}

    \begin{table*}
        [!t]
        \caption{Performance of large-scale and compact VLMs under zero-shot (ZS) setting
        on BRAR, DR, and MetaDent benchmarks. Best result across all models is in \textbf{bold},
        second-best is \underline{underlined}.}
        \label{tab:zs}
        \centering
        \small
        \setlength{\tabcolsep}{3pt}
        \begin{tabular}{llccccccc}
            \toprule \multirow{2}{*}{}                               & \multirow{2}{*}{Model} & \multicolumn{2}{c}{BRAR} & \multicolumn{2}{c}{DR} & \multicolumn{3}{c}{MetaDent} \\
            \cmidrule(lr){3-4} \cmidrule(lr){5-6} \cmidrule(lr){7-9} &                        & Acc ($\uparrow$)  & F1 ($\uparrow$)   & F1 ($\uparrow$)   & Acc ($\uparrow$) & VQA ($\uparrow$)             & Cap ($\uparrow$)         & Cls ($\uparrow$)  \\
            \midrule \multirow{7}{*}{\shortstack{Large\\VLMs\\(ZS)}}        & Lingshu-32B            & 0.490             & \underline{0.392} & \textbf{0.380}    & 0.082 & \underline{0.660}            & 0.196                    & \underline{0.325} \\
                                                                     & MedMO-8B-Next          & 0.255             & 0.191             & 0.210             & 0.041 & 0.387                        & 0.135                    & 0.056             \\
                                                                     & Qwen2.5-VL-7B          & 0.255             & 0.160             & 0.217             & 0.014 & 0.620                        & 0.200                    & 0.184             \\
                                                                     & gemini-2.5-flash       & 0.456             & \textbf{0.411}    & 0.317             & \underline{0.178} & \textbf{0.729}               & 0.194                    & \textbf{0.412}    \\
                                                                     & gpt-4o-mini            & \underline{0.557} & 0.239             & 0.123             & 0.041 & 0.600                        & \textbf{0.242}           & 0.286             \\
            \cmidrule{2-9}                                           & \textit{Mean}          & \textit{0.403}    & \textit{0.279}    & \textit{0.249}    & \textit{0.071} & \textit{0.599}               & \textit{0.193}           & \textit{0.253}    \\
            \midrule \multirow{11}{*}{\shortstack{Compact\\VLMs\\(ZS)}}       & Qwen3-VL-4B            & 0.436             & 0.369             & 0.100             & 0.014 & 0.620                        & \underline{0.232}        & 0.298             \\
                                                                     & gemma-4-E4B-it         & \underline{0.557} & 0.239             & 0.270             & \textbf{0.219} & 0.580                        & 0.168                    & 0.254             \\
                                                                     & InternVL2\_5-4B        & 0.295             & 0.195             & 0.078             & 0.027 & 0.520                        & 0.166                    & 0.116             \\
                                                                     & medgemma-4b-it         & 0.443             & 0.330             & \underline{0.350} & 0.068 & \underline{0.660}            & 0.168                    & 0.139             \\
                                                                     & paligemma2-3b-mix-448  & 0.086             & 0.053             & 0.000             & 0.000 & 0.000                        & 0.000                    & 0.000             \\
                                                                     & SmolVLM2-2.2B          & 0.544             & 0.292             & 0.000             & 0.000 & 0.000                        & 0.100                    & 0.158             \\
                                                                     & InternVL3\_5-2B        & 0.537             & 0.259             & 0.287             & 0.082 & 0.560                        & 0.173                    & 0.187             \\
                                                                     & gemma-4-E2B-it         & \textbf{0.564}    & 0.240             & 0.000             & 0.000 & 0.640                        & 0.140                    & 0.190             \\
                                                                     & InternVL3\_5-1B        & 0.228             & 0.219             & 0.317             & \underline{0.178} & 0.520                        & 0.149                    & 0.146             \\
            \cmidrule{2-9}                                           & \textit{Mean}          & \textit{0.410}    & \textit{0.244}    & \textit{0.156}    & \textit{0.065} & \textit{0.456}               & \textit{0.144}           & \textit{0.154}    \\
            \bottomrule
        \end{tabular}
    \end{table*}

    \begin{table*}[!t]
        \caption{Performance under 1-shot (FS-1) and 2-shot (FS-2) few-shot settings on BRAR, DR, and MetaDent benchmarks. Best result across all models is in \textbf{bold}, second-best is \underline{underlined}.}
        \label{tab:fs}
        \centering
        \small
        \setlength{\tabcolsep}{3pt}
        \begin{tabular}{llccccccc}
            \toprule \multirow{2}{*}{} & \multirow{2}{*}{Model}
            & \multicolumn{2}{c}{BRAR} & \multicolumn{2}{c}{DR}
            & \multicolumn{3}{c}{MetaDent} \\
            \cmidrule(lr){3-4} \cmidrule(lr){5-6} \cmidrule(lr){7-9}
            & & Acc ($\uparrow$) & F1 ($\uparrow$)
            & F1 ($\uparrow$) & Acc ($\uparrow$)
            & VQA ($\uparrow$) & Cap ($\uparrow$) & Cls ($\uparrow$) \\
            \midrule
            \multirow{7}{*}{\shortstack{Large\\VLMs\\(FS-1)}}
            & Lingshu-32B      & 0.195 & 0.126 & 0.511 & 0.041 & \underline{0.780} & 0.160 & 0.105 \\
            & MedMO-8B-Next    & 0.174 & 0.099 & \underline{0.532} & 0.014 & 0.560 & 0.160 & 0.026 \\
            & Qwen2.5-VL-7B    & 0.309 & 0.237 & 0.173 & 0.027 & 0.580 & 0.210 & 0.187 \\
            & gemini-2.5-flash & 0.383 & \textbf{0.338} & 0.457 & \textbf{0.278} & \textbf{0.796} & 0.207 & \textbf{0.404} \\
            & gpt-4o-mini      & 0.544 & \underline{0.314} & 0.503 & \underline{0.151} & 0.560 & \textbf{0.224} & \underline{0.295} \\
            \cmidrule{2-9}
            & \textit{Mean}    & \textit{0.321} & \textit{0.223}
            & \textit{0.435} & \textit{0.102} & \textit{0.655} & \textit{0.192} & \textit{0.203} \\
            \midrule
            \multirow{11}{*}{\shortstack{Compact\\VLMs\\(FS-1)}}
            & Qwen3-VL-4B           & 0.181 & 0.117 & 0.445 & 0.096 & 0.660 & \underline{0.222} & 0.282 \\
            & gemma-4-E4B-it        & 0.510 & 0.260 & 0.182 & 0.041 & 0.640 & 0.190 & 0.249 \\
            & InternVL2\_5-4B       & 0.195 & 0.133 & 0.324 & 0.041 & 0.580 & 0.167 & 0.097 \\
            & medgemma-4b-it        & 0.174 & 0.100 & 0.319 & 0.123 & 0.700 & 0.207 & 0.134 \\
            & paligemma2-3b-mix-448 & 0.170 & 0.097 & 0.000 & 0.000 & 0.000 & 0.000 & 0.000 \\
            & SmolVLM2-2.2B         & 0.248 & 0.185 & \textbf{0.533} & 0.014 & 0.511 & 0.145 & 0.153 \\
            & InternVL3\_5-2B       & \underline{0.557} & 0.239 & 0.083 & 0.055 & 0.660 & 0.213 & 0.136 \\
            & gemma-4-E2B-it        & \textbf{0.570} & 0.266 & 0.121 & 0.055 & 0.580 & 0.183 & 0.198 \\
            & InternVL3\_5-1B       & 0.389 & 0.251 & 0.284 & 0.110 & 0.520 & 0.204 & 0.026 \\
            \cmidrule{2-9}
            & \textit{Mean}    & \textit{0.333} & \textit{0.183}
            & \textit{0.255} & \textit{0.059} & \textit{0.539} & \textit{0.170} & \textit{0.142} \\
            \midrule[0.08em]
            \multirow{7}{*}{\shortstack{Large\\VLMs\\(FS-2)}}
            & Lingshu-32B      & 0.356 & 0.299 & 0.495 & 0.014 & \textbf{0.760} & \textbf{0.246} & 0.110 \\
            & MedMO-8B-Next    & \textbf{0.564} & 0.240 & 0.514 & 0.027 & 0.540 & 0.160 & 0.028 \\
            & Qwen2.5-VL-7B    & 0.315 & 0.255 & 0.136 & 0.000 & 0.540 & 0.195 & 0.119 \\
            & gemini-2.5-flash & 0.383 & 0.342 & \textbf{0.547} & 0.208 & \underline{0.755} & 0.145 & \textbf{0.316} \\
            & gpt-4o-mini      & 0.389 & \textbf{0.375} & \underline{0.533} & 0.082 & 0.520 & \underline{0.239} & 0.268 \\
            \cmidrule{2-9}
            & \textit{Mean}    & \textit{0.401} & \textit{0.302}
            & \textit{0.445} & \textit{0.066} & \textit{0.623} & \textit{0.197} & \textit{0.168} \\
            \midrule
            \multirow{11}{*}{\shortstack{Compact\\VLMs\\(FS-2)}}
            & Qwen3-VL-4B           & 0.315 & 0.310 & 0.391 & 0.082 & 0.560 & 0.228 & \underline{0.289} \\
            & gemma-4-E4B-it        & 0.403 & 0.286 & 0.176 & 0.041 & 0.620 & 0.184 & 0.230 \\
            & InternVL2\_5-4B       & \textbf{0.564} & 0.240 & 0.227 & 0.041 & 0.540 & 0.219 & 0.084 \\
            & medgemma-4b-it        & \underline{0.557} & \underline{0.347} & 0.361 & \textbf{0.260} & 0.660 & 0.186 & 0.124 \\
            & paligemma2-3b-mix-448 & 0.174 & 0.099 & 0.000 & 0.000 & 0.000 & 0.000 & 0.000 \\
            & SmolVLM2-2.2B         & \underline{0.557} & 0.239 & 0.516 & 0.027 & 0.500 & 0.112 & 0.160 \\
            & InternVL3\_5-2B       & 0.523 & 0.276 & 0.309 & 0.041 & 0.600 & 0.198 & 0.196 \\
            & gemma-4-E2B-it        & \textbf{0.564} & 0.240 & 0.318 & 0.041 & 0.600 & 0.168 & 0.158 \\
            & InternVL3\_5-1B       & \textbf{0.564} & 0.241 & 0.323 & \underline{0.219} & 0.580 & 0.184 & 0.026 \\
            \cmidrule{2-9}
            & \textit{Mean}    & \textit{0.469} & \textit{0.253}
            & \textit{0.291} & \textit{0.084} & \textit{0.518} & \textit{0.164} & \textit{0.141} \\
            \bottomrule
        \end{tabular}
    \end{table*}

    Table~\ref{tab:zs} reports zero-shot performance across all 14~models.

    \textbf{Zero-shot performance.} Under zero-shot evaluation, no model is consistently best across all metric columns. Gemini-2.5-Flash achieves the best BRAR F1 (0.411), MetaDent VQA accuracy (0.729), and classification F1 (0.412), while Lingshu-32B leads DR F1 (0.380). Among compact models, performance is fragmented across different tasks: gemma-4-E4B-it leads DR Acc (0.219), gemma-4-E2B-it leads BRAR Acc (0.564), illustrating that no single model dominates across all tasks. These results indicate that zero-shot evaluation alone is not a reliable basis for dental deployment.

    \textbf{Few-shot performance.} Table~\ref{tab:fs} reports 1-shot and 2-shot performance separately. Moving from ZS to 1-shot, DR classification improves moderately (large-VLM mean: 0.435 vs.\ 0.249 ZS), suggesting that in-context exemplars help anchor lesion category boundaries. MetaDent VQA also benefits (compact-VLM mean: 0.539 vs.\ 0.456 ZS). However, moving from 1-shot to 2-shot shows mixed or declining gains on most metrics, consistent with HealthSLM-Bench~\cite{healthslmbench}, which observed diminishing returns beyond 1--3 shots in health domains. Notably, several models report identical BRAR accuracy under the few-shot settings (e.g., 0.564 in FS-2); confusion-matrix inspection reveals that these models collapse to predicting only the majority class (Grade~2, which constitutes 84 of 149 test samples), yielding an accuracy equal to the majority-class prior rather than reflecting genuine discriminative ability. Detailed metric definitions are provided in Appendix~\ref{app:eval-metrics}.

    \subsection{Compact Adaptation Under a Uniform LoRA Budget}
    \label{sec:sft}

    Table~\ref{tab:sft} presents LoRA fine-tuning results across all 12 open-weight models. We focus on best-in-class comparisons rather than tier-level averages, since the large-VLM tier is heterogeneous (two of three models exhibit format collapse on classification; see Section~\ref{sec:failure}).

    \textbf{Effect of dental-domain adaptation.} Under LoRA instruction tuning, compact VLMs improve substantially on most tasks. InternVL3.5-2B (2B) achieves the best BRAR accuracy (0.651 vs.\ 0.584) and the best BRAR F1 (0.633 vs.\ 0.497) among all open-weight models. On DR classification, gemma-4-E4B-it (4B) achieves the best Macro~F1 (0.759), surpassing Lingshu-32B (0.651). On MetaDent, InternVL3.5-2B achieves the best captioning score (0.286 vs.\ Lingshu-32B's 0.244). The largest adaptation gain appears in DR classification: zero-shot Macro~F1 averages 0.189 across all 14 models; after LoRA, gemma-4-E4B-it reaches 0.759 F1, demonstrating that multi-label lesion classification can be reliably learned with lightweight domain adaptation.

    \textbf{Comparison with larger open-weight models.} Under the uniform low-cost adaptation budget ($r{=}16$, 3 epochs), compact models can match or exceed much larger ones on the majority of tasks. InternVL3.5-2B matches or outperforms larger open-weight models (7B--32B) on 4 of 5 primary metrics. \rev{We include closed-source APIs only as non-adapted cloud baselines, since their weights are unavailable for LoRA adaptation or on-device deployment; therefore, these results should not be read as an intrinsic capability comparison.} However, we cannot rule out that a higher-rank or longer-schedule configuration would close this gap for large models.

    \textbf{Role of medical pre-training.} medgemma-4b-it and paligemma2-3b-mix-448, both pre-trained on biomedical corpora, do not consistently outperform general-purpose models of comparable size after SFT. paligemma2-3b-mix-448 collapses to near-zero on BRAR after LoRA (Acc/F1~=~0). medgemma-4b-it is competitive on BRAR (F1 0.439) but lags behind gemma-4-E4B-it and InternVL3.5-2B on DR and captioning tasks. One possible explanation is that broad biomedical pre-training may not sufficiently cover dental-specific visual patterns, acquisition geometry, and structured output conventions, whereas LoRA adaptation directly aligns the model with the curated dental task distribution.

    \begin{table*}
        [!t]
        \caption{Performance of large-scale and compact VLMs under \textbf{instruction tuning (LoRA)}
        on BRAR, DR, and MetaDent benchmarks. Best result across all models is in \textbf{bold},
        second-best is \underline{underlined}.}
        \label{tab:sft}
        \centering
        \small
        \setlength{\tabcolsep}{3pt}
        \begin{tabular}{llccccccc}
            \toprule \multirow{2}{*}{}                               & \multirow{2}{*}{Model} & \multicolumn{2}{c}{BRAR} & \multicolumn{2}{c}{DR} & \multicolumn{3}{c}{MetaDent} \\
            \cmidrule(lr){3-4} \cmidrule(lr){5-6} \cmidrule(lr){7-9} &                        & Acc ($\uparrow$)  & F1 ($\uparrow$)   & F1 ($\uparrow$)   & Acc ($\uparrow$) & VQA ($\uparrow$)             & Cap ($\uparrow$)         & Cls ($\uparrow$)  \\
            \midrule \multirow{5}{*}{\shortstack{Large\\VLMs\\(LoRA)}}      & Lingshu-32B            & 0.584             & 0.497             & 0.651             & 0.507 & \textbf{0.920}               & 0.244                    & 0.000 \\
                                                                     & MedMO-8B-Next          & 0.174             & 0.099             & 0.219             & 0.288 & 0.820                        & 0.252                    & 0.000             \\
                                                                     & Qwen2.5-VL-7B          & 0.564             & 0.421             & 0.605             & 0.521 & 0.840                        & 0.237                    & 0.101    \\
            \cmidrule{2-9}                                           & \textit{Mean}          & \textit{0.441}    & \textit{0.339}    & \textit{0.492}    & \textit{0.439} & \textit{0.860}               & \textit{0.244}           & \textit{0.034}    \\
            \midrule \multirow{11}{*}{\shortstack{Compact\\VLMs\\(LoRA)}}     & Qwen3-VL-4B            & 0.570             & \underline{0.549} & 0.636             & 0.603 & 0.820                        & 0.226                    & 0.116             \\
                                                                     & gemma-4-E4B-it         & 0.550             & 0.423             & \textbf{0.759}    & \textbf{0.712} & \underline{0.880}            & 0.262                    & \textbf{0.343}    \\
                                                                     & InternVL2\_5-4B        & 0.523             & 0.521             & 0.536             & 0.521 & 0.820                        & 0.271                    & 0.287             \\
                                                                     & medgemma-4b-it         & \underline{0.624} & 0.439             & 0.561             & 0.479 & 0.780                        & 0.261                    & 0.246             \\
                                                                     & paligemma2-3b-mix-448  & 0.000             & 0.000             & 0.000             & 0.000 & 0.556                        & 0.094                    & 0.168             \\
                                                                     & SmolVLM2-2.2B          & 0.537             & 0.508             & 0.395             & 0.260 & 0.780                        & \underline{0.277}        & 0.331             \\
                                                                     & InternVL3\_5-2B        & \textbf{0.651}    & \textbf{0.633}    & \underline{0.732} & \underline{0.699} & 0.820                        & \textbf{0.286}           & 0.316             \\
                                                                     & gemma-4-E2B-it         & 0.174             & 0.099             & 0.687             & 0.644 & 0.820                        & 0.225                    & \underline{0.335}             \\
                                                                     & InternVL3\_5-1B        & 0.517             & 0.490             & 0.730             & \textbf{0.712} & 0.800                        & 0.275                    & 0.284    \\
            \cmidrule{2-9}                                           & \textit{Mean}          & \textit{0.461}    & \textit{0.407}    & \textit{0.560}    & \textit{0.514} & \textit{0.786}               & \textit{0.242}           & \textit{0.255}    \\
            \bottomrule
        \end{tabular}
    \end{table*}

    \begin{table*}
        [!t]
        \centering
        \caption{On-device deployment efficiency of VLMs on iPhone~17~Pro
        (MetaDent dataset, $N{=}30$ samples).}
        \label{tab:deployment_efficiency} \small
        \setlength{\tabcolsep}{5pt}
        \renewcommand{\arraystretch}{1.1}
        \resizebox{\textwidth}{!}{%
        \begin{tabular}{llrrrrrrr}
            \toprule Dataset                   & Model             & TTFT\,(s)\,$\downarrow$ & ITPS\,(t/s)\,$\uparrow$ & OET\,(s)\,$\downarrow$ & OTPS\,(t/s)\,$\uparrow$ & Total\,(s)\,$\downarrow$ & CPU\,(\%)\,$\downarrow$ & RAM\,(GB)\,$\downarrow$ \\
            \midrule \multirow{3}{*}{MetaDent} & Pocket-Dentist-2B & \textbf{0.76} & \textbf{447.95} & \textbf{3.55} & \textbf{30.58} & \textbf{4.31}  & 64.06          & \textbf{2.62} \\
                                               & InternVL2.5-4B    & 1.64          & 202.80           & 4.94          & 21.09          & 6.58           & 51.22          & 3.08          \\
                                               & Qwen2.5-VL-7B     & 4.88          & 68.98            & 16.25         & 9.17           & 21.13          & \textbf{42.85} & 6.03          \\
            \bottomrule
        \end{tabular}%
        }
    \end{table*}

    \subsection{On-Device Deployment Feasibility}
    \label{sec:deployment}

    Table~\ref{tab:deployment_efficiency} reports the latency and resource utilization of top-performing LoRA-tuned VLMs deployed on an iPhone~17~Pro (12\,GB Unified Memory) via on-device Metal-accelerated inference (cf.\ Appendix~\ref{app:deployment}). This deployment experiment is not intended as a clinical validation study; it tests whether benchmark-competitive compact VLMs can satisfy the memory and latency constraints of local mobile inference.

    The deployment results reveal a clear accuracy--efficiency trade-off. Pocket-Dentist-2B (InternVL3.5-2B + LoRA) reduces end-to-end latency from 21.13\,s for the 7B baseline to 4.31s per sample (a 4.9$\times$ reduction), while reducing memory use from 6.03\,GB to 2.62\,GB (2.3$\times$ reduction). This places the benchmark-competitive compact model well within the memory budget of the test device and substantially lowers the cost of local inference. The 4B mid-point (InternVL2.5-4B) achieves 6.58\,s latency with 3.08\,GB RAM, providing an intermediate operating point on the accuracy--efficiency frontier.

    \textbf{Token efficiency.} On-device profiling reveals that all three models consume identical prompt budgets ($\sim$346 input tokens averaged across tasks), as they share the same structured prompts. Output token counts are also comparable, averaging 97, 93, and 110 tokens per sample for Pocket-Dentist-2B, InternVL2.5-4B, and Qwen2.5-VL-7B, respectively, indicating that the compact model produces answers of comparable length. The efficiency gap, therefore, arises from throughput: Pocket-Dentist-2B generates output at 30.6\,tokens/s versus Qwen2.5-VL-7B's 9.2\,tokens/s, a 3.3$\times$ improvement in token-generation efficiency. All throughput figures (ITPS, OTPS) are reported as the mean of per-sample ratios rather than the ratio of aggregated totals (see Appendix~\ref{app:deployment} for details). This demonstrates that compact LoRA-adapted models do not sacrifice answer length for speed; they produce comparable output while requiring substantially fewer compute resources per token.

    Beyond raw performance, this on-device deployment approach offers practical advantages relevant to resource-constrained settings: (i)~patient privacy, as dental images containing identifiable intraoral features remain on the device, mitigating data-exposure risks inherent in cloud-based inference; (ii)~network independence, as the system functions in offline settings common to rural clinics and mobile dental outreach programs. Combined with the benchmark results showing that \rev{the SFT-adapted Pocket-Dentist-2B is competitive with zero-/few-shot closed-source APIs in this deployment-scenario comparison} (Section~\ref{sec:sft}), these results demonstrate the feasibility of local on-device inference for dental image understanding (see Figure~\ref{fig:deployment} in Appendix~\ref{app:deployment} for the on-device evaluation interface).

    \subsection{Output Reliability and Failure Modes}
    \label{sec:failure}

    Deployment-oriented evaluation must account for output reliability,
    not just task accuracy. We observe two distinct failure modes
    across our benchmark.

    \textbf{Format collapse.} Among the large-scale models, Lingshu-32B and MedMO-8B-Next exhibit clear format-collapse behavior after LoRA adaptation. Both produce classification F1~=~0 because they generate free-form text instead of the expected JSON array of category IDs. Under the uniform adaptation budget ($r{=}16$, 3~epochs), these larger models exhibited structured-output instability, suggesting that their stronger text-generation priors may require higher adaptation capacity or constrained decoding to produce valid structured outputs. This pattern echoes the class-imbalance collapse documented in HealthSLM-Bench~\cite{healthslmbench}. Notably, this format collapse depresses the tier-level mean comparisons in Table~\ref{tab:sft}: the large-VLM classification mean (0.034) reflects failure modes rather than capability, which is why we report best-in-class comparisons in Section~\ref{sec:sft}.

    \textbf{Architectural incompatibility.} Under zero-shot and few-shot settings, paligemma2-3b-mix-448 cannot process MetaDent's multi-turn prompt format (all MetaDent and DR metrics reported as 0). While it begins producing valid formats after LoRA adaptation, it collapses to BRAR Acc/F1 = 0, indicating fundamental mismatches between its fixed-resolution architecture and the task requirements.

    These failure cases highlight that efficiency-aware QA evaluation must account for output reliability alongside raw accuracy. For efficient deployment, these failures also represent wasted compute: models that produce unusable outputs consume the same memory and latency budget as models that produce correct answers.

    \subsection{Threats to Validity}
    \label{sec:threats}

    We identify three threats that readers should weigh when
    interpreting our results.

    \textbf{Patient-level separation.} This cannot be fully verified in MetaDent.
    The dataset combines images from a university dental clinic and
    curated web sources, with an estimated $\sim$171 unique patients.
    Because no formal patient-level identifier is available, we
    cannot fully rule out patient overlap between training and test
    splits. BRAR, by contrast, provides verified one-patient-per-image
    provenance.

    \textbf{Clinical provenance.} The DR benchmark is a community-
    contributed Kaggle collection without published clinical
    provenance, patient demographics, or
    inter-annotator agreement statistics. Results on this dataset
    should be interpreted as a proof-of-concept for VLM classification
    rather than clinical evidence.

    \textbf{Deployment hardware.} All on-device measurements are collected on a single iPhone~17~Pro,
    chosen to accommodate the 7B baseline model (6.03\,GB RAM). The
    actual deployed model (Pocket-Dentist-2B, 2.62\,GB) has
    substantially lower hardware requirements and is expected to run
    on mid-range devices with 4\,GB+ RAM, including budget Android
    smartphones commonly available in resource-constrained settings.
    Nevertheless, no cross-platform latency or thermal measurements
    have been conducted, and performance on lower-spec hardware
    remains to be validated.


    \section{Limitations}
    \label{sec:limitations}
    \rev{We report single-run point estimates without confidence intervals or significance tests. Because several evaluation sets are small (DR $n{=}73$, BRAR $n{=}149$, $N{=}30$ on-device), some differences may fall within sampling or run-to-run noise and are indicative rather than definitive. We also do not characterize LoRA seed variance. Bootstrap confidence intervals and multi-seed analysis are left to future work.}


    \section{Conclusion}
    \label{sec:conclusion}

    We present Pocket-Dentist, an efficiency-aware multimodal QA benchmark and evaluation pipeline for dental VLMs. The benchmark standardizes three dental datasets across panoramic radiographs and intraoral photographs into a unified dental QA evaluation comprising five task types and seven metrics. On this benchmark, zero-shot and few-shot QA performance is fragmented across tasks, whereas dental-domain LoRA adaptation enables compact VLMs, particularly InternVL3.5-2B, to \rev{become competitive with substantially larger open-weight models on most primary metrics} under a uniform low-cost adaptation budget. Pocket-Dentist-2B achieves 4.31s per-sample inference on an iPhone~17~Pro with fully local computation, demonstrating a 4.9$\times$ latency and 2.3$\times$ memory reduction relative to the 7B baseline while maintaining competitive QA accuracy. The model brings preliminary oral-health screening to 4GB+ budget smartphones. The benchmark does not establish clinical efficacy or safety; it provides an empirical foundation for future work on clinical validation, efficient multimodal QA in broader healthcare domains, and cross-institution deployment studies.

    \newpage
    \section*{Impact Statement}
    This paper presents work to advance Machine Learning. Our benchmark and on-device deployment pipeline support preliminary oral-health screening in resource-constrained settings, performing local inference to preserve patient privacy. The system is a research tool and does not constitute a clinical diagnostic device. There are many potential societal consequences of our work, none which we feel must be specifically highlighted here.

    
    \bibliographystyle{icml2026}
    \bibliography{refs}


    \newpage
    \appendix
    \onecolumn

    \section{Prompt Templates and Output Schemas}
    \label{app:prompt-details}

    This section details the structured input--output formats used
    for each benchmark.

    \textbf{BRAR.}
    The prompt presents a panoramic radiograph together with
    patient-level metadata (age, gender, missing teeth, implants,
    residual roots, functional tooth count) and requests a single
    severity grade. The expected output is a JSON object
    \texttt{\{"grade": \textit{k}\}} where $k \in \{1,2,3\}$.
    Because models produce outputs in heterogeneous formats
    (e.g., plain digits, JSON objects, markdown code blocks, or
    embedded text), we apply a five-level deterministic fallback
    parser: (1)~pure digit match, (2)~JSON parsing, (3)~markdown
    code-block extraction, (4)~regex pattern matching, and
    (5)~first-digit heuristic.

    \textbf{MetaDent.}
    Three task-specific output schemas are used.
    \textit{VQA}: \texttt{\{"answer": ..., "reason": ...\}}.
    \textit{Classification}: a JSON array of detected conditions,
    e.g., \texttt{[\{"id": "C1", "name": "Dental caries",
    "evidence": "..."\}]}.
    \textit{Captioning}:
    \texttt{\{"description": "..."\}}.

    \textbf{DR.}
    The prompt describes the four supported finding categories
    (Cavity, Fillings, Impacted Tooth, Implant) with radiographic
    definitions. The model is asked to identify which categories
    are present in the panoramic radiograph.
    The expected output is a structured JSON listing the detected
    categories, e.g., \texttt{\{"objects":[\{"label":"Cavity"\},
    \{"label":"Fillings"\}]\}}.


    %
    \section{Evaluation Metrics}
    \label{app:eval-metrics}

    This section provides formal definitions and clinical rationale for all
    evaluation metrics used in our benchmark tables.
    Arrows in column headers indicate optimization direction:
    ($\uparrow$)~higher is better; ($\downarrow$)~lower is better.

    \subsection{Task-Level Metrics ($\uparrow$)}

    BRAR Accuracy / F1 ($\uparrow$) evaluate three-class classification performance (Grade~1/2/3) for periodontal bone
    resorption grading in panoramic radiographs. Accuracy measures overall
    correctness, while Macro~F1
    accounts for class imbalance across resorption
    severity grades. In clinical practice, misgrading bone resorption
    can lead to delayed intervention in progressive
    periodontal disease, making higher recall---and consequently higher
    F1---clinically imperative.

    DR Macro F1 / Accuracy ($\uparrow$) evaluate image-level multi-label classification
    across four lesion categories (Cavity, Fillings, Impacted Tooth, Implant)
    in panoramic radiographs. Macro~F1 is the unweighted mean of per-class F1
    scores, ensuring balanced evaluation across categories regardless of
    prevalence. Accuracy measures the proportion of images whose predicted
    category set exactly matches the ground-truth set. In clinical practice,
    correctly identifying the types of lesions present in a radiograph is a
    prerequisite for appropriate treatment planning.

    VQA Accuracy ($\uparrow$) denotes the proportion of correctly answered visual questions, computed as exact-match
    accuracy against clinician-verified reference answers. In dental diagnostics,
    a higher VQA accuracy directly translates to more reliable triage
    recommendations---an incorrect answer (e.g., misidentifying a lesion as
    ``normal'') could delay critical treatment.

    Captioning BERTScore F1 ($\uparrow$) calculates a semantic similarity score between generated captions and reference
    descriptions, computed via contextual embeddings. Unlike surface-level
    $n$-gram overlap (BLEU/ROUGE), BERTScore captures paraphrase equivalence,
    which is essential in clinical reporting where lexical variation is high but
    semantic fidelity is paramount. Higher scores indicate that the model's
    narrative more faithfully conveys the clinical findings present in the
    radiograph.
    \rev{We compute BERTScore with \texttt{bert-score} v0.3.13 using
    \texttt{roberta-large} layer~17, \texttt{lang="en"}, \texttt{idf=False}, and
    \texttt{rescale\_with\_baseline=True}. Thus, the reported F1 scores
    ($0.09$--$0.29$) are rescaled BERTScore values relative to a precomputed
    random baseline, not raw BERTScore values; the monotonic rescaling is
    applied uniformly to all models and preserves their ranking.}

    Classification F1 ($\uparrow$) represents the harmonic mean of precision and recall across 18 dental condition
    categories. F1 is preferred over raw accuracy because the MetaDent label
    distribution is heavily long-tailed: a model that predicts only the majority
    class achieves high accuracy but clinically dangerous low recall on rare
    pathologies (e.g., periapical abscess). Higher F1 ensures balanced
    sensitivity across both common and rare conditions.

    \subsection{Deployment Efficiency Metrics}

    TTFT ($\downarrow$), OET ($\downarrow$), and Total Time ($\downarrow$) are latency metrics measured in seconds. In a chairside clinical workflow, the
    dentist awaits the model's response while the patient is present; excessive
    latency disrupts the consultation flow and reduces practitioner trust.
    Lower values directly improve clinical usability. TTFT captures perceived
    responsiveness (time to first token); OET and Total Time capture end-to-end
    throughput for complete diagnostic output.

    ITPS ($\uparrow$) and OTPS ($\uparrow$) are throughput metrics measured in tokens per second. Higher input throughput
    (ITPS) enables faster processing of multimodal prompts containing
    high-resolution dental images; higher output throughput (OTPS) accelerates
    the generation of structured diagnostic responses. Together, they
    characterize the computational efficiency of the inference engine on
    resource-constrained mobile hardware.

    CPU ($\downarrow$) and RAM ($\downarrow$) denote resource utilization metrics. Lower CPU usage preserves battery life and
    reduces thermal throttling during sustained clinical sessions. Lower RAM
    footprint determines deployment feasibility on devices with limited unified
    memory (e.g., 12\,GB on iPhone 17 Pro) and leaves headroom for the host
    application's UI and camera pipeline.

    \section{Deployment Setup and Metrics}
    \label{app:deployment}

    \subsection{Hardware and Software Environment}

    \begin{itemize}
        \item Device: iPhone 17 Pro, iOS 26.3.1, Apple A19 Pro SoC, 12\,GB
            Unified Memory.

        \item Inference Engine: \texttt{llama.cpp} via \texttt{LocalLLMClient}
            Swift package, mapping Metal API to GPU-accelerated inference.

        \item Quantization: All models converted to GGUF format with Q4\_K\_M
            (4-bit) quantization.

        \item Architecture: Pure on-device inference (Route A)---no
            network communication; models are sideloaded to the app's Documents directory
            via Finder and loaded entirely into Unified Memory.
    \end{itemize}

    \begin{wrapfigure}{r}{0.4\columnwidth}
        \centering
        \vspace{-12pt}
        \includegraphics[width=0.24\columnwidth]{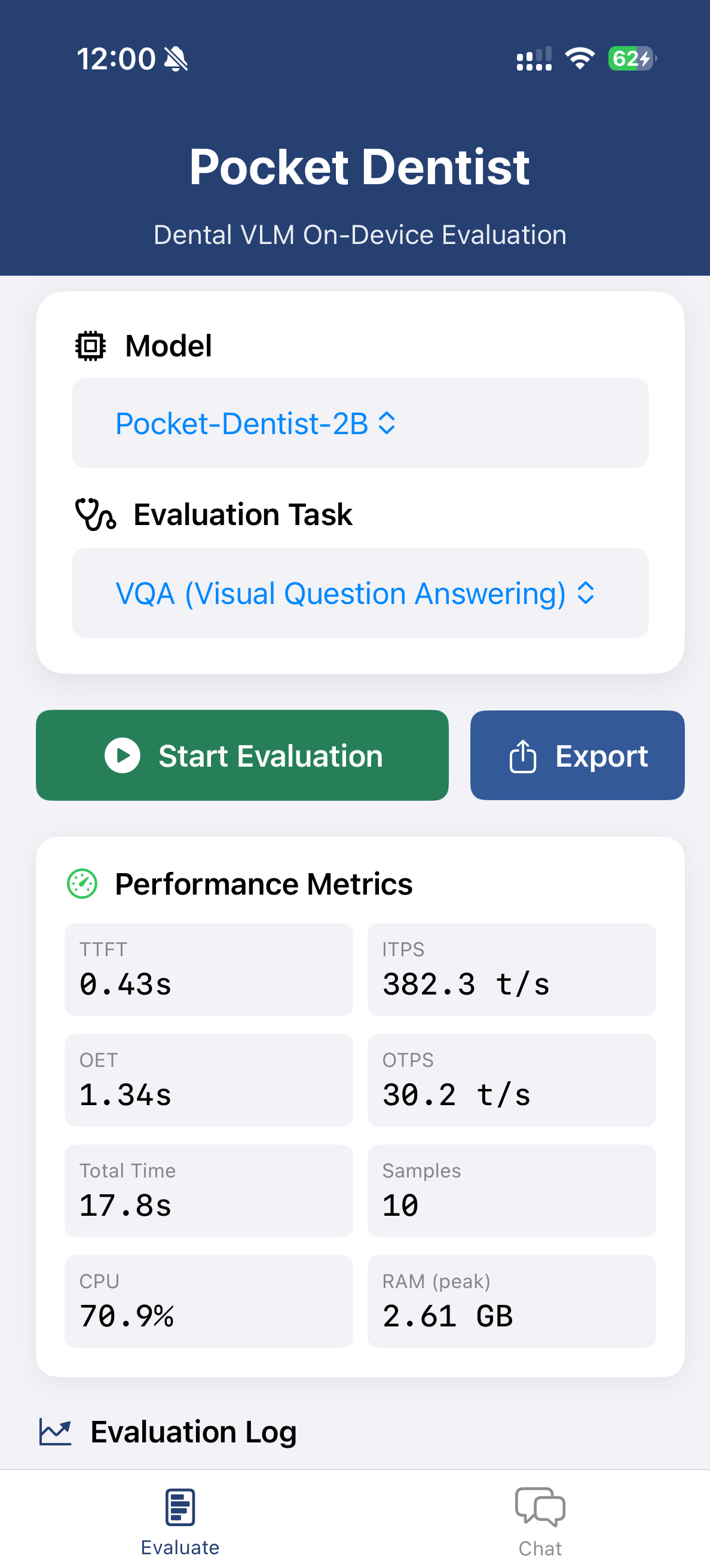}
        \caption{Pocket-Dentist iOS app running Pocket-Dentist-2B
        on an iPhone~17~Pro.}
        \label{fig:deployment}
        \vspace{-10pt}
    \end{wrapfigure}

    \subsection{Efficiency Metrics}

    Following MobileAIBench~\cite{mobileaibench}, we adopt:
    \begin{itemize}
        \item TTFT (s): Time-to-first-token---latency from prompt
            submission to first output token.

        \item ITPS (t/s): Input tokens per second---prompt processing
            throughput.

        \item OTPS (t/s): Output tokens per second---generation
            throughput.

        \item OET (s): Output evaluation time---wall-clock time for
            complete response generation.

        \item Total Time (s): End-to-end per-sample latency (TTFT + OET).

        \item CPU (\%): Average CPU utilization during inference.

        \item RAM (GB): Peak memory allocation during model inference.
    \end{itemize}

    \subsection{Aggregation}

    Latency metrics are averaged over $N = 30$ samples (10 per task $\times$ 3
    tasks). Following HealthSLM-Bench methodology, MetaDent is treated as a single
    data pool without task-level stratification for efficiency measurement, as
    latency metrics reflect computational engine characteristics rather than
    task-specific semantics. CPU utilization is the per-second average load; RAM
    is the peak allocation.

\end{document}